\documentclass{article}
\usepackage{ijcai25}

\usepackage{times}
\usepackage{soul}
\usepackage{url}
\usepackage[hidelinks]{hyperref}
\usepackage[utf8]{inputenc}
\usepackage[small]{caption}
\usepackage{graphicx}
\usepackage{amsmath}
\usepackage{amsthm}
\usepackage{multirow}
\usepackage{booktabs}
\usepackage{algorithm}
\usepackage{algorithmic}
\usepackage[switch, mathlines]{lineno}  
\usepackage{amssymb} 

\title{Multi-modal Deepfake Detection and Localization with FPN-Transformer}

\author{
    Chende Zheng\and 
    Ruiqi Suo\and 
    Zhoulin Ji\and 
    Jingyi Deng\and
    Fangbin Yi \and  
    Chenhao Lin\thanks{Corresponding Author} \and
    Chao Shen
    \affiliations
    Xi’an Jiaotong University 
    \emails
    \{zhengchende, srain22, 310449, yi446320, misscc320\}@stu.xjtu.edu.cn,
    linchenhao@xjtu.edu.cn,
    chaoshen@mail.xjtu.edu.cn
}

\begin{document}
\maketitle

\begin{abstract}

The rapid advancement of generative adversarial networks (GANs) and diffusion models has enabled the creation of highly realistic deepfake content, posing significant threats to digital trust across audio-visual domains. While unimodal detection methods have shown progress in identifying synthetic media, their inability to leverage cross-modal correlations and precisely localize forged segments limits their practicality against sophisticated, fine-grained manipulations. 
To address this, we introduce a multi-modal deepfake detection and localization framework based on a Feature Pyramid-Transformer (FPN-Transformer), addressing critical gaps in cross-modal generalization and temporal boundary regression. The proposed approach utilizes pre-trained self-supervised models (WavLM for audio, CLIP for video) to extract hierarchical temporal features. A multi-scale feature pyramid is constructed through R-TLM blocks with localized attention mechanisms, enabling joint analysis of cross-context temporal dependencies. The dual-branch prediction head simultaneously predicts forgery probabilities and refines temporal offsets of manipulated segments, achieving frame-level localization precision. 
We evaluate our approach on the test set of the IJCAI'25 DDL-AV benchmark, showing a good performance with a final score of 0.7535 for cross-modal deepfake detection and localization in challenging environments. Experimental results confirm the effectiveness of our approach and provide a novel way for generalized deepfake detection. 
Our code is available at \href{https://github.com/Zig-HS/MM-DDL}{\textit{https://github.com/Zig-HS/MM-DDL}}.

\end{abstract}


\section{Introduction}
The rapid advancement of generative adversarial networks (GANs) and diffusion models has revolutionized audio-visual synthesis technologies, enabling the creation of highly realistic deepfake content that is increasingly indistinguishable from authentic media. While these breakthroughs have driven innovations in entertainment, education, and human-computer interaction, they have also introduced severe security threats. Malicious actors exploit synthetic speech synthesis (e.g., voice conversion and text-to-speech systems) and facial manipulation techniques (e.g., face swapping and attribute editing) to generate deceptive content for identity impersonation, misinformation dissemination, and social engineering attacks. This dual-use nature of generative AI underscores the urgent need for robust deepfake detection and localization frameworks capable of safeguarding digital trust in real-world applications.  

Despite significant progress in unimodal deepfake detection (e.g., audio spoofing analysis or video forgery identification), existing methods face two critical limitations. First, most approaches operate in isolated modalities, failing to leverage cross-modal correlations that could enhance detection accuracy and generalization. For instance, audio-visual consistency cues, such as lip-sync alignment or speaker identity coherence, are often ignored in single-modality pipelines. Second, while temporal localization of forged segments (e.g., identifying manipulated intervals in a video or audio clip) is crucial for forensic analysis, current solutions struggle with partial manipulations (e.g., spliced speech segments or localized facial edits) due to their reliance on simplistic binary classification or rigid post-processing heuristics. These shortcomings hinder practical deployment, particularly against evolving deepfake techniques that exploit multimodal and fine-grained editing strategies.  

To address these challenges, we propose a unified multi-modal deepfake detection and localization framework based on Feature Pyramid-Transformer (FPN-Transformer).
We leverage pretrained self-supervised models (WavLM~\cite{chen2022wavlm} for audio, CLIP~\cite{radford2021learning} for video) to extract hierarchical temporal features, followed by masked differential convolution for local context modeling. Subsequently, we construct a multi-scale feature pyramid using R-TLM~\cite{sun2021transformer} blocks with 1D downsampling convolutions to jointly analyze audio-visual temporal dependencies through local attention mechanisms. Finally, we employ a dual-branch prediction head to simultaneously predict forgery probabilities and precisely localize the start/end offsets of manipulated segments.
To validate the effectiveness of our framework, we conduct extensive experiments on the IJCAI'25 DDL-AV benchmark. Experimental results demonstrate that our proposed method achieves superior detection and localization performance in a comprehensive evaluation.
Our main contributions can be summarized as follows:

\begin{itemize}
    \item We propose a general-purpose temporal data forgery detection model for multimodal deepfake localization, addressing the challenge that existing detection frameworks struggle to handle multimodal forged data.
    \item We conduct extensive experiments on the IJCAI'25 dataset to validate the effectiveness of our framework with a final score of 0.7535, providing novel solutions toward achieving generalized forgery detection.
\end{itemize}



\section{Related Works}
\subsection{Synthetic Audio Generation}
The rapid advancement of deep learning and generative models has significantly propelled the development of speech synthesis technologies. Early speech synthesis methods primarily relied on parametric synthesis (e.g., formant synthesis \cite{styger1994formant}, linear predictive coding (LPC) \cite{o1988linear}) and concatenative synthesis (e.g., unit selection synthesis \cite{hunt1996unit}), yet these approaches exhibited notable limitations in naturalness and flexibility. 

Currently, the field of speech synthesis is dominated by generative adversarial networks (GANs) and diffusion models, which can produce highly natural speech, often indistinguishable from human voices. For instance, models such as VITS \cite{kim2021conditional} and YourTTS \cite{casanova2022yourtts} integrate variational inference and speaker adaptation techniques, enabling high-quality multi-speaker synthesis with limited data. Additionally, diffusion-based probabilistic models (e.g., DiffWave \cite{kong2020diffwave}, WaveGrad \cite{chen2020wavegrad}) generate speech through iterative noise addition and denoising processes, further enhancing synthesis quality.

Recent advancements in large language models (LLMs) have spurred the development of autoregressive speech synthesis. Models like VALL-E \cite{wang2023neural} leverage text-driven audio synthesis to achieve more natural speech prosody. However, the progress in speech synthesis has also raised security concerns regarding audio deepfakes. Malicious actors may exploit voice conversion (VC) and text-to-speech (TTS) technologies for identity impersonation, posing threats to social trust and privacy security. Consequently, developing highly robust fake speech detection methods has become an urgent need for both academia and industry.

\subsection{Synthetic Video Generation}


Synthetic Video Generation is an emerging technological field that aims to fabricate facial expressions and movements of target individuals through AI techniques, thereby producing deceptive content. So far, Synthetic Video Generation can be categorized into three types: Face Swapping, Facial Reenactment, and Attribute Manipulation.

Face-swapping methods aim to replace a target face with a source face while preserving the original context and identity consistency.~\cite{ding2020swapped} introduced a method that performs reliably across various scenes and maintains identity fidelity. However, it suffers from imperfect face-region blending, often producing boundary artifacts and being sensitive to changes in lighting or pose.
\cite{zendran2021swapping} combined VAE and GAN architectures to better recover facial details and improve latent space controllability, but the training process tends to be unstable, and the method struggles with facial expression alignment.

Facial reenactment methods generate new images or videos that transfer expressions or movements from one face to another. LE-GAN by~\cite{hu2023face} uses a Laplacian pyramid to enhance facial detail and produce more realistic expressions. However, it can suffer distortions under rapid motion or large pose variations.
B-GAN~\cite{liu2020generating} exploits frequency domain modeling to enhance texture realism and reduce GAN artifacts, yet it's sensitive to occlusions and may produce blurry edges. 

Attribute manipulation methods modify facial features such as age, expression, or hairstyle.~\cite{li2023sc} proposed SC-GAN, which embeds semantic directions into the latent space for fine-grained control and natural output. However, when editing multiple attributes simultaneously, the model can exhibit semantic conflicts, and undefined regions in the latent space may lead to unnatural results. GANprintR by \cite{neves2020ganprintr} focuses on fingerprint removal from GAN-generated images to enhance their stealthiness. While effective against detection, it often sacrifices texture quality. \cite{manjula2022deep} combined StarGAN, C-GAN, and VAE for flexible multi-attribute control, but the integration is complex and prone to unstable training. 

\subsection{Synthetic Audio Detection}
The field of fake speech detection originated with the use of various digital signal processing algorithms to analyze anomalies in audio signals for distinguishing genuine from fake audio. Early researchers conducted foundational studies through statistical modeling and acoustic feature analysis \cite{malik2010audio}. With the release of increasingly large-scale datasets such as ASVspoof \cite{wang2020asvspoof}, research in this domain has deepened significantly. Subsequently, deep learning-based approaches have gained prominence, achieving remarkable results across multiple dimensions.

 Early approaches to forgery detection relied primarily on digital signal processing techniques without involving neural network-based methods. \cite{malik2010audio} proposed leveraging spectral decay characteristics to detect forged speech, achieving reasonable performance while maintaining strong interpretability. With the advent of neural networks, machine learning-based techniques were increasingly adopted by researchers as countermeasures for forged speech detection. \cite{alegre2012spoofing} introduced an SVM-based robust forgery detection strategy, significantly outperforming traditional digital signal processing methods. Subsequently, deep learning techniques were widely applied to forged speech detection. 


Recently, the generalization capability of audio deepfake detection has garnered increasing attention from researchers, as it determines whether a detection method can perform effectively in real-world scenarios or merely achieve seemingly high metrics on limited academic datasets. \cite{muller2022does} discusses the impact of different factors on generalization and highlights that many existing approaches perform poorly on real-world data, indicating that some solutions in the research community have become overly tailored to specific datasets, thereby losing generalizability and practical usability.  \cite{chen2020generalization} improves the loss function by employing IMCL, a loss supervision mechanism that enhances feature separation, instead of relying solely on simple binary cross-entropy loss. Some researchers have also adopted single-domain generalization techniques to ensure that the feature space exhibits relatively strong generalization \cite{xie2023single}. These methods are typically trained and fine-tuned exclusively on large-scale real-world data , enabling them to achieve robust detection results against various unseen attacks. 

Simple binary classification has become insufficient for handling certain complex scenarios, such as partially manipulated audio, where only specific segments of a sample are generated by speech synthesis algorithms. These manipulated segments often distort the original semantic meaning of the audio, posing significant risks. \cite{yi2022add} introduced the concept of utterance-level fake speech detection, demonstrating through their challenge dataset that baseline models employing traditional binary classification methods perform poorly when detecting samples containing partially substituted words or phrases generated by speech synthesis algorithms. Consequently, the detection and localization of partially manipulated speech has emerged as a critical research focus. 

\subsection{Synthetic Video Detection}




Deepfake detection methods focus primarily on three types of features: temporal features, spatial features, and frequency features. Most mainstream detection approaches are based on one of these domains or integrate multiple feature types to improve robustness.

Temporal-based deepfake detection methods focus primarily on the temporal consistency between video frames and abnormal behavioral patterns. These approaches model dynamic changes across frame sequences to identify forgeries. Since fake videos often fail to reproduce genuine human physiological behaviors, such as blinking, head movements, and gaze direction,~\cite{li2018ictu} detect anomalies in blinking frequency to determine video authenticity.
~\cite{yang2019exposing} analyze inconsistencies in head pose by comparing estimates from all facial landmarks with those from central regions.~\cite{peng2024deepfakes} focus on interframe gaze angle variation and propose a spatio-temporal feature fusion module that combines gaze dynamics, spatial attributes, and texture features for classification. Frame-to-frame discontinuities also serve as key indicators of manipulation.~\cite{zheng2024breaking} reveal that existing detectors suffer from semantic artifacts across diverse scenes and propose a patch-shuffling strategy to break these artifacts for generalized detection.~\cite{yin2023dynamic} design a multiscale spatiotemporal aggregation module to model interframe inconsistencies, while \cite{choi2024exploiting} observe fluctuations in latent variable styles between frames and develop a style attention module accordingly. 

Spatial-based detection methods  often analyzing single image frames to identify forged artifacts such as texture anomalies, blending boundaries, and lighting inconsistencies.  For example, RECCE~\cite{cao2022end} utilizes reconstruction consistency to detect shadows learned from real images, while LGrad~\cite{tan2023learning} converts images into gradient representations using a pre-trained transformation network to amplify subtle artifacts.~\cite{ba2024exposing} go beyond isolated regions and integrate information from multiple nonoverlapping areas to detect more global inconsistencies.~\cite{miao2024mixture} further introduce a Mixture-of-Noises Module to enhance RGB features with noise traces, improving localization accuracy in multi-face manipulation detection. Other studies emphasize differences between facial and non-facial regions or leverage fine-grained texture cues~\cite{chai2020makes}~\cite{nirkin2021deepfake}.

Frequency-based detection methods transform visual information from the spatial or temporal domain into the frequency domain and analyze periodic signal characteristics to reveal manipulation traces.F3-Net~\cite{qian2020thinking} proposes a dual-branch framework: one branch uses frequency-aware image decomposition (FAD) to learn fine-grained forgery patterns, while the other extracts local frequency statistics (LFS) for semantic-level analysis. HFI-Net\cite{miao2022hierarchical} incorporates global-local interaction modules to explore multi-level frequency artifacts and further enhance detection performance. FreqNet~\cite{tan2024frequency} focuses on high-frequency components of images, combining them with a frequency-domain learning module to extract source-independent features and improve generalizability. Additionally,~\cite{miao2023multi} leverage multi-spectral class centers to suppress semantic information and enhance localization capabilities through frequency-aware features.

\section{Methodology}

\begin{figure*}[!t]
    \centering
    \includegraphics[width=1\textwidth]{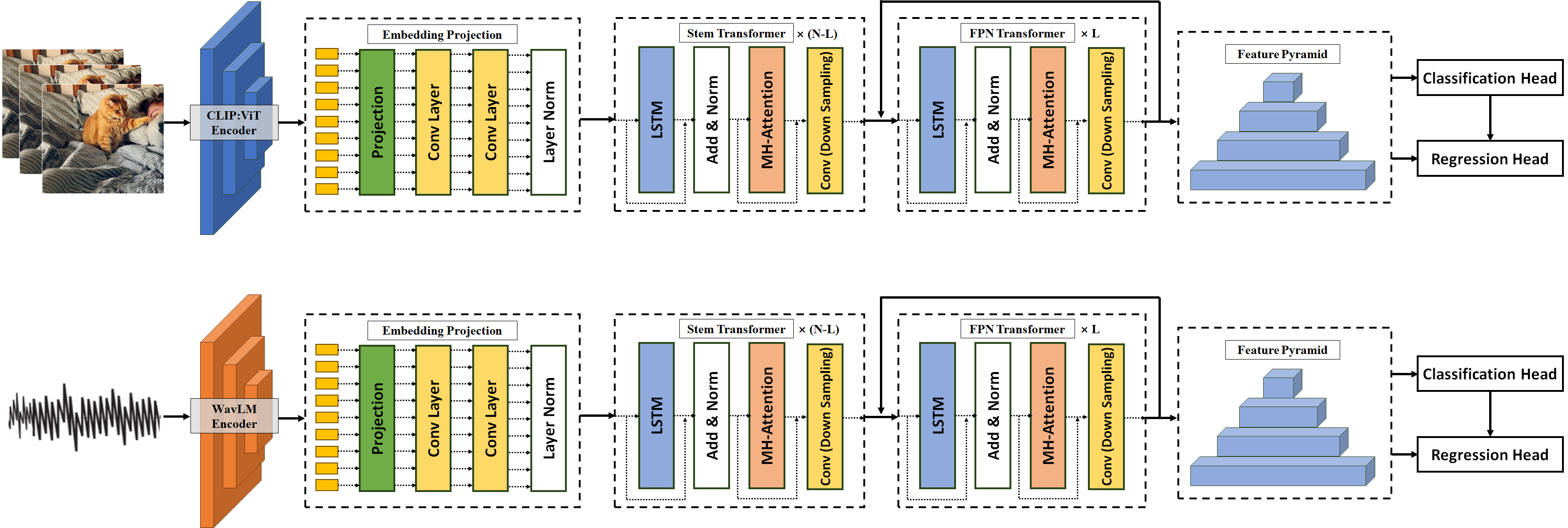}
        \caption{Framework of proposed dual FPN-Transformer detection method. First, we employ pre-trained self-supervised models (WavLM/CLIP:ViT) to encode input temporal data into feature embeddings. Then, these embeddings are processed through an encoder composed of Transformer blocks to obtain the feature pyramid. And finally, prediction heads jointly analyze temporal features to predict forgery boundaries. We train two models separately for audios and videos, and combine the output results.}
    \label{fig:pipeline}
\end{figure*}

Existing deepfake detection solutions are incapable of achieving generalizable cross-modal detection and localization. Based on our previous unimodal work~\cite{ji2024speech}, our approach enables the detection and localization of multimodal temporal forgeries. Notably, rather than simultaneously processing audio-visual modalities, our approach treats cross-modal data as unified temporal feature sequences. Figure~\ref{fig:pipeline} illustrates the proposed dual FPN-Transformer detection architecture, which comprises three key components: temporal feature embedding and projection module, FPN-Transformer backbone module, and prediction heads. First, during the feature embedding stage, we employ pre-trained self-supervised models (WavLM/CLIP) to encode input temporal data (audio/video) into feature embeddings, generating temporal feature sequences. Subsequently, these embeddings are processed through an encoder composed of lightweight Transformer blocks to obtain the feature pyramid. And finally, classification and regression heads jointly analyze temporal features to predict forgery boundaries at each time step, specifically estimating start times and offset distances of forged segments for precise localization. The following sections will elaborate on these components in detail.

\subsection{Feature Embedding and Projection}
\paragraph{Problem Definition.} 
Traditional audio or video forgery detection and localization methods exclusively focus on single modal inputs, inherently limiting their cross-modal generalizability. To address this limitation, we propose a unified problem formulation for multimodal temporal data (audio/video). Specifically, given a temporal input sequence $ X = \{x_1, \dots, x_T\} $, our objective is to generate the corresponding output sequence $ Y = \{y_1, \dots, y_N\} $ through a mapping function $ f: X \rightarrow Y $. Each element $ y_n = (p_n, d^s_n, d^e_n) $ of the output $ Y $ denotes one potential forged segment, where  $ p_n $ denotes the forgery probability of the $ n $-th segment, $ d^s_n $ denotes the start time offset relative to the input sequence, and $ d^e_n $ denotes the end time offset relative to the input sequence.  

Notably, for genuine (non-forged) sequences, the expected output $ Y $ should be an empty set. This formulation enables precise temporal localization of forged segments while maintaining cross-modal consistency through unified temporal feature representations.

Considering the distinct temporal characteristics across modalities (e.g., audio data exhibits dense temporal resolution with sampling rates reaching kilohertz levels, while video data presents sparse temporal structure with frame rates limited to hertz levels), we implement a unified feature embedding framework through temporal encoding. Specifically, the input sequence $ X = \{x_1, \dots, x_T\} $ is encoded by encoder $ e $ into feature representations $ Z = \{z_1, \dots, z_M\} $. For each feature representation $ z_i $, $ f_L $ denotes the temporal sequence length corresponding to $ z_i $ and $ f_S $ denotes the temporal offset between consecutive feature representations. Consequently, the length $ M $ of feature sequence $ Z $ is given by the following equation:
\begin{align}
M = \left\lfloor \frac{T - f_L}{f_S} \right\rfloor + 1  
\end{align}
where $ \lfloor \cdot \rfloor $ denotes integer flooring. 

To facilitate batch processing, variable-length sequences are standardized through padding/truncation to a maximum length $ M_{max} $, with masking mechanisms ensuring valid temporal context propagation.

Inspired by previous research~\cite{ojha2023towards}, we employ pre-trained self-supervised models with frozen weights as encoder $ e $. Specifically, WavLM-LARGE~\cite{chen2022wavlm} is adopted for audio data while CLIP:ViT-B/16~\cite{radford2021learning} is adopted for video data. Compared to conventional feature calculating approaches (e.g., LFCC, MFCC, image DFT), these self-supervised models have been exposed to massive training data, enabling superior capability in capturing low-level features critical for differentiating genuine and forged content~\cite{yang2021superb}. The architectural design leverages this property to detect subtle discrepancies inherent in generative artifacts.  


Subsequently, we employ a set of masked differential convolutional networks to implement feature projection, which facilitates positional embedding integration and effectively captures local temporal context. Specifically, for a given feature embedding $ z \in Z $, the output of the masked 1D differential convolution at timestamp $ t_0 $ is formulated as:  

\begin{align}
\text{MDC}(t_0) =  \theta \cdot \left(-z(t_0) \cdot \sum_{t_n \in D} w(t_n)\right) \\
+ \sum_{t_n \in D} w(t_n) \cdot z(t_0 + t_n)
\end{align}

where $ t_0 $ denotes the current timestamp, $ t_n $ denotes the enumerated timestamps in offset set $ D $, $ w(t_n) $ denotes the learnable convolutional weights, and $ \theta \in [0,1] $ is a hyperparameter balancing the contribution between intensity-level and gradient-level information  

\subsection{FPN-Transformer Architecture}
We employ $ N $ layers of R-TLM blocks~\cite{sun2021transformer} to perform deep feature encoding. Compared to standard Transformer architectures, R-TLM incorporates additional LSTM and Fusion layers to explicitly model cross-context representation interactions. The multi-head self-attention (MSA) layer in R-TLM integrates temporal context across the sequence. Notably, we apply localized attention masking to constrain computations within sliding windows, motivated by two factors: (1) forged segments exhibit localized temporal characteristics, and (2) this design significantly reduces computational complexity.  

To capture hierarchical temporal features at multiple scales for constructing a feature pyramid, we integrate R-TLM with strided 1D depthwise convolutions. Specifically, we introduce a strided depthwise 1D convolution after each MSA layer. By aggregating outputs from multi-level R-TLM structures, we obtain a hierarchical feature pyramid $ \mathcal{F} = \{F^{(1)},...,F^{(L)}\} $ with $ L $ levels. 

A critical component involves temporal alignment between feature sequence timestamps $ \tau $ and original input timestamps $ t $. Given a virtual timestamp $ \tau_{i} $ at the $ i $-th pyramid level and its cumulative stride factor $ s_i $, we map $ \tau_{i} $ to the corresponding physical timestamp $ t $ in the raw input domain through:  
\begin{align}
t = \left\lfloor \frac{s_i}{2} \right\rfloor + \tau_{i} \cdot s_i
\end{align}


\begin{figure*}[!t]
    \centering
    \includegraphics[width=1\textwidth]{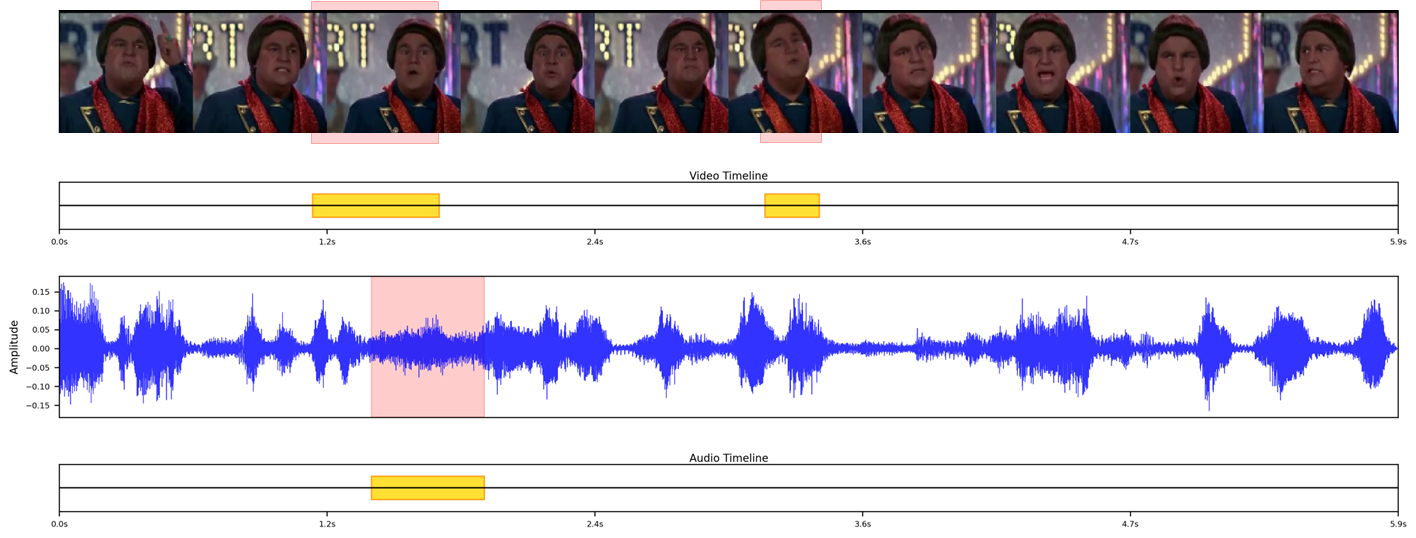}
        \caption{Visualization results of our proposed method. Red represents forged segments, and yellow represents our predicted results. Our method can accurately predict the presence of forged video and audio segments in the samples for both video and audio modalities.}
    \label{fig:visualization}
\end{figure*}

\subsection{Prediction Head} 
We employ a dual-branch prediction head to decode the feature pyramid $ \mathcal{F} $ into the desired output $ Y $. The decoder consists of:  

\paragraph{Classification Head.} Given feature pyramid $ \mathcal{F} $, the classification head evaluates all $ L $ pyramid levels at each timestamp $ t $ to predict the forgery probability $ p(t) $. This is implemented through lightweight 1D convolutional networks attached to each pyramid level, with parameters shared across levels. Specifically, the classification network comprises 3 convolutional layers (kernel size=3), layer normalization (applied to first two layers), and ReLU activation. A final sigmoid function is applied to output dimensions to produce probabilistic forgery predictions.  


\paragraph{Regression Head.} Distinct from the classification head, the regression head predicts temporal boundaries only when timestamp $ t $ lies within forged segments (During inference, we utilize the classification head's output to determine whether a timestamp lies within forged segments). For each pyramid level, we predefine an output regression range to model the start offset $ d^s_t $ and end offset $ d^e_t $. The regression head employs 1D convolutional networks with ReLU activation to ensure precise distance estimation. Specifically, the most probable forgery span $ [s_t, e_t] $ corresponding to timestamp $ t $ is determined by:  
\begin{align}
c_t = \mathop{\text{argmax}} p(c_t),\quad s_t = t - d^s_t, \quad e_t = t - d^e_t
\end{align}

\subsection{Loss Function}
Our prediction task involves dual objectives: (1) binary classification of forgery probability at each timestamp $ t $, and (2) temporal boundary regression for forged segments (start/end offsets). We design a composite loss function combining two components:  

\paragraph{Classification Loss.}
We employ focal loss~\cite{lin2017focal} to address class imbalance between forged and genuine segments. For timestamp $ t $, the classification loss ($ \mathcal{L}_{\text{cls}} $) is formulated as:  
\begin{align}
\mathcal{L}_{\text{cls}} = -\frac{1}{T^+} \sum_{t=1}^T 1(t \in \Omega^+) \cdot \left[ \log(p_t) + \gamma \cdot (1-p_t)^\alpha \right]
\end{align} 
where $ p_t $ denotes predicted forgery probability, $ \Omega^+ $ represents forged regions, $ T^+ = |\Omega^+| $ is the total number of positive samples, and $ \alpha $ and $ \gamma $ are hyperparameters balancing hard mining effects.

\paragraph{Regression Loss.}  
For timestamps $ t \in \Omega^+ $, we minimize DIoU loss ($ \mathcal{L}_{\text{reg}} $)~\cite{zheng2020distance} between predicted boundaries $ \hat{s}_t, \hat{e}_t $ and ground-truth $ s_t^*, e_t^* $:  
\begin{align}
\mathcal{L}_{\text{reg}} = \frac{1}{T^+} \sum_{t \in \Omega^+} \left( 1 - \text{DIoU}(\hat{s}_t, \hat{e}_t; s_t^*, e_t^*) \right)
\end{align}
\paragraph{Final Loss.}
The overall objective combines both components:   
\begin{align}
\mathcal{L}_{\text{total}} = \frac{1}{T^+} \sum_{t=1}^T \left[ \lambda \mathcal{L}_{\text{cls}} + \mathcal{I}(t \in \Omega^+) \mathcal{L}_{\text{reg}}(t) \right]
\end{align}
where $ \lambda \in [0,1] $ is the balancing ratio between classification and localization tasks, and $ \mathcal{I}(t \in \Omega^+) $ is an indicator function (1 if timestamp $ t $ lies in forged regions $ \Omega^+ $, 0 otherwise).

\begin{table*}[!t]
    \centering
    \begin{tabular}{ccccc}
    \toprule
    \multicolumn{2}{c}{Training strategy} & \multicolumn{2}{c}{Feature Embedding}& \multirow{2}{*}{Final Score} \\ 
    \cmidrule(r){1-2} \cmidrule(r){3-4}
    Initial Learning Rate & Epochs & Audio & Video &~ \\
    \midrule
        $1 \times 10^{-3}$ & 3 & wavLM & CLIP & 0.7535  \\
        $1 \times 10^{-3}$ & 6 & wavLM & CLIP & 0.7501  \\
        $1 \times 10^{-3}$ & 15 & wavLM & CLIP & 0.6590  \\ 
        $1 \times 10^{-3}$ & 36 & wavLM & CLIP & 0.6174  \\ 
        $1 \times 10^{-3}$ & 60 & wavLM & CLIP & 0.6144  \\
        $1 \times 10^{-3}$ & 95 & wavLM & CLIP & 0.6000  \\ 
        $1 \times 10^{-3}$ & 6 & wav2vec & XCLIP & 0.7361  \\
        $1 \times 10^{-3}$ & 60 & wav2vec & XCLIP & 0.5644  \\
        $1 \times 10^{-3}$ & 95 & wavLM & XCLIP & 0.5873  \\ 
        $1 \times 10^{-3}$ & 95 & wav2vec & XCLIP & 0.5798  \\ \hline
        $3 \times 10^{-4}$ & 2 & wavLM & CLIP & 0.5821  \\
        $3 \times 10^{-4}$ & 5 & wavLM & CLIP & 0.7164  \\ 
        $3 \times 10^{-4}$ & 6 & wavLM & CLIP & 0.7218  \\ 
        $3 \times 10^{-4}$ & 7 & wavLM & CLIP & 0.7218  \\ 
        $3 \times 10^{-4}$ & 8 & wavLM & CLIP & 0.7340  \\ 
        $3 \times 10^{-4}$ & 9 & wavLM & CLIP & 0.7252  \\ 
        $3 \times 10^{-4}$ & 10 & wavLM & CLIP & 0.7201  \\ 
        $3 \times 10^{-4}$ & 11 & wavLM & CLIP & 0.7256  \\ 
        $3 \times 10^{-4}$ & 12 & wavLM & CLIP & 0.7227  \\ 
        $3 \times 10^{-4}$ & 13 & wavLM & CLIP & 0.7182  \\ 
        $3 \times 10^{-4}$ & 14 & wavLM & CLIP & 0.7126  \\ 
        $3 \times 10^{-4}$ & 15 & wavLM & CLIP & 0.7084  \\ 
    \bottomrule
    \end{tabular}
    \caption{Comparison experiments of our proposed method under varying experimental conditions, including self-supervised feature extractors, training epochs, and initial learning rates. 
    }
    \label{tab:results}
\end{table*}

\section{Experiments}



\subsection{Implementation Details}

\paragraph{Preprocessing.} 
Audio data are resampled to 16 kHz and processed using WavLM-LARGE to extract 1024-dimensional feature vectors at 20 ms intervals (50 FPS). Video data undergoes frame extraction at 25 FPS, followed by resizing to 224×224 pixels and normalization. Per-frame feature extraction employs CLIP:ViT-B/16, generating 768-dimensional embeddings. All implementations utilize PyTorch on NVIDIA L40 GPUs.  

\paragraph{Training.} 
We adopt the AdamW optimizer with mini-batch processing, incorporating a 5-epoch warmup phase and cosine decay for learning rate scheduling. The initial learning rate is $1 \times 10^{-3}$ and the weight decay is $1 \times 10^{-2}$.
Variable-length sequences are standardized through padding/truncation to a maximum length of 1024, with masking mechanisms ensuring valid temporal context propagation. 
The number of R-TLM blocks $N$ is set to 6, and the number of FPN levels $L$ is set to 5.
$ \theta $ is set to 0.6 and the balancing ratio $ \lambda $ of the loss is set to 0.01. 
The model is trained for fixed epochs (up to 95) with a batch size of 64, and separate models are trained for audio and video modalities.


\paragraph{Inference.}
During inference, the complete sequence is input to the model with a batch size of 1. Non-Maximum Suppression (NMS)~\cite{neubeck2006efficient} is applied to refine predictions by eliminating highly overlapping and inefficient instances, yielding the final forged segment outputs. For unimodal temporal data, the maximum forgery confidence among all predicted segments is treated as the overall confidence score for the entire sequence. The final outputs are obtained by containing both audio and video modalities, where the higher confidence score between the two modalities is selected as the final sample-level forgery confidence, while the union of predicted audio and video forged segments forms the complete output set of forged regions.  

\paragraph{Dataset.}
We evaluate our framework on the IJCAI'25 DDL-AV dataset~\cite{miao2025ddl,zhang2024inclusion,zhang2024mfms} for model training and testing. The dataset comprises 200k video samples in the training set, 20k in the validation set, and 111k in the test set. These samples cover 9 deepfake audio forgery methods and 18 deepfake video forgery techniques, ensuring comprehensive evaluation across diverse generation paradigms.  

\paragraph{Metrics.} 
To comprehensively assess our method's performance across diverse scenarios, we adopt multi-dimensional metrics following the IJCAI'25 Workshop on Deepfake Detection and Localization dataset (DDL-AV dataset). Detection performance is quantified using Area Under Curve (AUC), while localization quality is measured via Average Precision (AP) and Average Recall (AR). The localization score is computed as:  
\begin{align}
\text{Score} & = \frac{1}{16} \text{AP@0.5} + \frac{2}{16} \text{AP@0.75} + \frac{2}{16} \text{AP@0.9} \\ 
             &\quad + \frac{3}{16} \text{AP@0.95} + \frac{1}{16} \text{AR@30} + \frac{2}{16} \text{AR@20}  \\
             &\quad + \frac{2}{16} \text{AR@10} + \frac{3}{16} \text{AR@5}
\end{align}

The final comprehensive performance metric is defined as:  
\begin{align}
\text{Final Score} = \frac{\text{AUC} + \text{Score}}{2}
\end{align}
This formulation balances detection accuracy and localization precision through weighted aggregation of threshold-specific metrics, ensuring robust evaluation of both global and fine-grained forgery identification capabilities.

\subsection{Experimental Results}




\paragraph{Visualization results.}
Figure~\ref{fig:visualization} illustrates the visualization of our method's prediction results on multimodal input data. The visualizations demonstrate that our framework effectively detects and localizes forged segments in both video and audio modalities, with clear temporal alignment between predicted boundaries and ground-truth forgery regions. This empirical evidence validates the cross-modal consistency and precision of our detection mechanism.  

\paragraph{Quantitative results.}
Table~\ref{tab:results} presents a comprehensive performance comparison of our proposed framework under varying experimental conditions, including self-supervised feature extractors, training epochs, and initial learning rates. Our optimal configuration achieves a final score of \textbf{0.7535}, significantly outperforming baseline variants.  

\begin{itemize}
    \item \textbf{Self-supervised feature extractors.}
    To evaluate modality-specific feature representation capabilities, we contrast two audio feature extractors (Wav2Vec~\cite{baevski2020wav2vec} vs. WavLM) and two video feature extractors (CLIP vs. XCLIP~\cite{ni2022expanding}). While WavLM and CLIP consistently yield superior performance (Table~\ref{tab:results}), the marginal drops observed in Wav2Vec/XCLIP configurations can be attributed to:  
    1) Wav2Vec has lower model capacity (768-dimensional features vs. WavLM's 1024D) and reduced contextual modeling capability due to its shallow transformer architecture;  
    2) Temporal misalignment arising from XCLIP's original pretraining on 8 FPS videos, which conflicts with DDL-AV's 25 FPS format. Despite incorporating temporal encoding mechanisms, XCLIP struggles to adapt to higher frame rate dynamics, leading to suboptimal boundary localization; 
    Notably, the performance gaps across extractors remain statistically insignificant, demonstrating the robustness of our framework against variations in low-level feature representations. This suggests that our hierarchical FPN-Transformer architecture successfully abstracts modality-specific discrepancies into unified temporal feature spaces.  
    
    \item \textbf{Training Depth and Learning Rate Analysis.}
    When training with $1 \times 10^{-3}$ initial learning rate, deeper training (higher epochs) correlates with reduced generalization performance on DDL-AV's test set, which contains forgery methods absent in training. This manifests as declining scores beyond 3 epochs (optimal point). Reducing the learning rate to $3 \times 10^{-4}$ reveals a non-monotonic generalization trend: performance improves initially but degrades at excessive depths. 
    This reveals a critical trade-off between training depth and generalization robustness. Our findings suggest that balancing model capacity with controlled training duration may mitigate performance drops on out-of-distribution forgery samples, offering a novel perspective for improving detection systems' adaptability to evolving deepfake techniques. 
    
\end{itemize} 

\section{Conclusion and Discussion}
This paper presents a novel multi-modal deepfake detection and localization framework based on FPN-Transformer, addressing critical limitations in cross-modal generalization and temporal boundary regression. 
By leveraging pre-trained self-supervised models (WavLM for audio, CLIP for video) to extract hierarchical temporal features, combined with a multi-scale feature pyramid constructed via R-TLM blocks and localized attention mechanisms, our approach achieves robust analysis of cross-context temporal dependencies. The dual-branch prediction head enables simultaneous forgery probability estimation and precise temporal offset refinement, achieving frame-level localization accuracy. 
Evaluated on the IJCAI’25 DDL-AV benchmark, our method attains a final score of 0.7535, demonstrating superior performance in detecting and localizing sophisticated, fine-grained manipulations in challenging environments. 

Our work advances the field by providing a unified solution for generalized deepfake detection, bridging the gap between unimodal approaches and complex real-world scenarios. Future research may explore dynamic adaptation to evolving generative techniques and further optimize computational efficiency for real-time applications.

\section*{Acknowledgments}
This research is supported by the National Key Research and Development Program of China (2023YFE0209800), the National Natural Science Foundation of China (T2341003, 62376210, 62161160337, 62132011, U24B20185, U21B2018, 62206217), the Shaanxi Province Key Industry Innovation Program (2023-ZDLGY-38).


\bibliographystyle{named}
\bibliography{ijcai25}

\end{document}